\documentclass[10pt,twocolumn,letterpaper]{article}

\usepackage{cvpr}
\usepackage{times}
\usepackage{epsfig}
\usepackage{graphicx}
\usepackage{amsmath}
\usepackage{amssymb}

\usepackage{tabularx}
\usepackage{booktabs}
\usepackage{multirow}

\usepackage[pagebackref=true,breaklinks=true,letterpaper=true,colorlinks,bookmarks=false]{hyperref}

\cvprfinalcopy 

\def\cvprPaperID{6393} 

\ifcvprfinal\fi
\begin{document}

\title{Mining YouTube - \\  A dataset for learning fine-grained action concepts from webly supervised video data}

\author{Hilde Kuehne, Ahsan Iqbal, Alexander Richard, Juergen Gall\\
University of Bonn, Germany\\
{\tt\small \{kuehne,iqbalm,richard,gall\}@iai.uni-bonn.de}}

\maketitle

\begin{abstract}

Action recognition is so far mainly focusing on the problem of classification of 
hand selected preclipped actions and reaching impressive results in this field. 
But with the performance even ceiling on current datasets, it 
also appears that the next steps in the field will have to go beyond this fully 
supervised classification. 
One way to overcome those problems is to move towards less restricted scenarios. In this context we present a 
large-scale real-world dataset designed to evaluate learning techniques for 
human action recognition beyond hand-crafted datasets. To this end 
we put the process of collecting data on its feet again and start with the 
annotation of a test set of 250 cooking videos. The training data is then 
gathered by searching for the respective annotated classes within the subtitles 
of freely available videos. The uniqueness of the dataset is attributed to the 
fact that the whole process of collecting the data and training does not involve 
any human intervention. To address the problem of semantic inconsistencies that arise with this kind of training data, we further propose a semantical hierarchical structure for the mined classes.

We benchmark the proposed dataset with respect to current features and architectures on the task of temporal alignment and show challenges in this field as well as the benefits of semantic models in this context.
\end{abstract}


\section{Introduction}

Action recognition has become a more and more important topic in the field of 
academic research as well as in industrial context. This is shown by the amount 
of publications and the diversity of research directions, as well as by the 
growing number of challenging datasets  in this 
field \cite{caba2015activitynet,carreira2017quo,zhou2017procnets,gu2017ava}. 
So far, most of these approaches rely on fully supervised training. In this 
case, for any action class to be trained, preclipped videos or a label with 
temporal information is needed. But the acquisition of such data is very time 
consuming as it needs some kind of human supervision, e.g. in form of Mechanical 
Turk workers as used for Charades~\cite{sigurdsson2016hollywood} or Kinetics 
\cite{carreira2017quo}. This puts a natural limit to the size of current 
benchmarks and makes it unlikely to ever have ImageNet scale benchmarks in 
action recognition with millions of samples. 
Additionally, those datasets are usually based on a hand-crafted class vocabulary 
based on easy to search categories, e.g. from sports, as authors need to cover 
many different scenarios and at the same time, identify unique distinguishable 
actions. This results not only in a certain class overlap between the different 
datasets, it also has the effect that many datasets have a high inter class 
variation, such as `riding horse' vs. `playing piano' and a low intra class 
variation, which lead to the fact that many actions can be distinguished by one 
frame only and puts up few incentives to e.g. address temporal structures within 
the video.
Another problem related to the focus on fully supervised learning is that, so far, 
techniques developed and fine-tuned on those datasets do not naturally transfer 
to applications in the wild, as obviously pre-trained actions classes do usually 
not overlap with the vocabulary needed for specific applications. Further, 
collecting hand-annotated training data for those scenarios is often not 
feasible considering the amount of data needed and the fact that realistic 
classes are not always easily searchable.
\begin{figure}[t]
    \centering
    \includegraphics[width=0.45\textwidth]{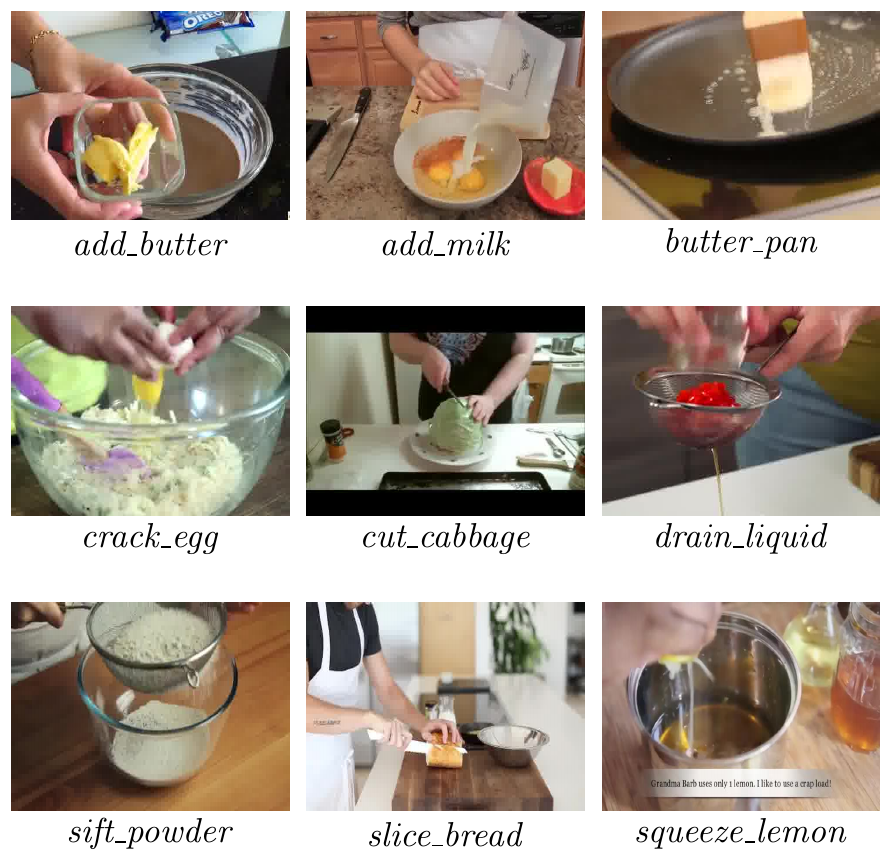}
    \caption{Examples of frames and related classes from YouTube cooking videos. 
Similar to \cite{malmaud15what}, we follow the idea of automatically mining 
large scale training data from videos and subtitles without the need for human 
intervention. }
    \label{fig:examples_classes}
\end{figure}
To address this issue, we propose a new benchmark addressing the problem of 
learning actions from real-live videos without human supervision. Our 
contribution is two-fold. First, we densely annotated a set of 250 cooking 
videos with several minutes duration each with realistic action labels as they 
occur in the instructions given in the video or in textual recipes. Examples for 
such occuring labels can be ``crack egg'', ``cut tomato'' or ``butter pan''. The 
resulting annotations comprise 512 action classes. Overall, we label 10,412 
samples within those videos. 
As the annotation of a training set for this amount of classes would be very 
time consuming, we want to avoid to compile training data by hand. Web-based 
crawling of the target classes might in this case become a suitable alternative, 
but, as we will show in the evaluation, it is not feasible for the here 
addressed target classes. 
We therefore address the problem of mining arbitrary action classes based on subtitle 
descriptions only and collected a set of training videos by searching class 
related attributes on YouTube using the transcripts of the audio stream to 
generate weak labels from the training examples. Overall we extract about $200k$ 
samples from 21,056 videos for training. Some examples of the mined clips and 
their related class labels are shown in Figure \ref{fig:examples_classes}. 
This approach follows and continues the idea of earlier works such as 
\cite{malmaud15what,laptev08learning,bojanowski2014weakly} and extends their 
work to large scale real live conditions. The goal of this paper is to make the 
whole process of gathering training samples and evaluating respective systems 
reproducible by providing all data necessary and as well as the respective fixed 
test set and hopefully leading to new ideas and insights in this field.

We evaluate the mined data with respect to current state-of-the-art features and 
architectures. It shows that even approaches that perform well on current 
curated datasets struggle with this kind of data. One problem here is that in case of uncontrolled mining, classes are no longer clearly separated by a human curator. Instead, we find several semantic inconsistencies like homonyms, polysemies, and other forms of inter-class relatedness. To address this problem we build, as second contribution, a sematical hierachical tree over all 512 action classes, with general classes such as ``chop it'' forming nodes close to the root and specific concepts such as ``cook bacon'' representing the leaves of the tree. Note that this hierarchy differs from most other common dataset hierarchies such as those for ActivityNet \cite{caba2015activitynet} or Kinetics \cite{carreira2017quo} as each node in the graph can be trained independently instead of only defining a hierachical chain for the leaf-nodes. It shows that the incorporation of semantically related classes can significantly improve the classification results in this case.  

\section{Related Work}

Action recognition has been a challenging topic for long and a lot of innovative 
approaches, mainly for the task of action classification 
\cite{laptev08learning,wang2013action,simonyan2014two},  have come up in the 
research community. But, obviously, we are still far away from the real-world 
task of learning arbitrary action classes from video data. One limitation here 
might be the lack of availability of real-world datasets that are just based on 
real random collections of videos. 

Apart from first generation datasets \cite{blank2005database,kth2004database} 
where actors were required to perform certain actions in controlled environment, 
current datasets such as HMDB \cite{kuehne2011hmdb}, UCF \cite{soomro2012ucf101} 
or the recently released Kinetics dataset \cite{carreira2017quo} are mainly 
acquired from web sources such as YouTube clips or movies with the aim to 
represent realistic scenarios from training and testing. Here, videos were 
usually first searched by predefined action-queries and later clipped and 
organized to capture the atomic actions or its repetitions. 
Other datasets such as Thumos \cite{thumos14}, MPI Cooking 
\cite{rohrbach2012database}, Breakfast\cite{kuehne2014breakfast} or the recently released Epic Kitchen dataset \cite{Damen2018EPICKITCHENS} focus on the 
labeling of one or more action segments in single long videos, \ie trying to 
temporally detect or segment predefined action classes within the video. 

The here proposed setting of mining actions from subtitles or scripts has also 
been considered. One of the first works in this field has been proposed by 
Laptev \etal \cite{laptev08learning} where the authors try to learn actions from 
movie scripts. Closer to the here followed approach is the scenario evaluated by 
Malmaud \etal \cite{malmaud15what}. Here, the authors also collected random 
cooking videos from YouTube and used a combination of subtitles, audio and 
vision to identify receipt steps in videos. However, there are still two main differences. 
\cite{malmaud15what} focuses on video with accompanying descriptive text \eg 
textual recipes as an additional cue for parsing and supervision. The problem 
here is that the evaluation is only done manually via Mechanical Turk workers 
which makes it hard to reproduce the results or compare the system to other 
methods. 
Another dataset that follows the idea of using subtitles has been proposed by 
Alayrac \etal \cite{alayrac2016unsupervised}, also using YouTube videos to 
automatically learn actions from instructional videos. Here, the main limitation 
is that all actions in the video have to follow a fix order, thus the order of 
the instructions has to be always the same. This limits the scope of possible 
training samples and still requires a human to pre-select suitable videos. 
Overall the dataset comprises five different tasks with a total of 150 videos 
and about 800,000 frames and is thus also significantly smaller than the here 
proposed one.

But not only datasets, also new techniques have evolved to address this kind of 
data. Arandjelovic and Zisserman \cite{arandjelovic2017looklistenlearn} explore 
the mapping between speech/audio-transcripts to the video for the task of audio 
visual classification. Kaufman \etal \cite{kaufman2016tessellation} approach the 
problem of semantic similarities in videos by using semantic transfer techniques 
from 2D image analysis. So far, those techniques are mainly benchmarked on fully 
supervised scenarios, but they are also applicable for less supervised scenarios 
as the here presented one.
Further, there are several works trying to adapt the idea of using web-crawled 
videos for different learning tasks. Hong \etal \cite{hong2017weakly} propose an 
approach for semantic segmentation, starting  with a set of weakly annotated 
images and extending the system in a second set to learn from web video data. 
Senner \etal \cite{Sener2017unsupervised} use collections of instructional 
videos from WikiHow to learn the semantic story line of different tasks. Similar 
to \cite{malmaud15what} and \cite{arandjelovic2017looklistenlearn}, they also 
feature a multi modal approach and learn visual atoms from object proposals and 
word frequencies to capture mid-level semantic concepts and construct high level 
activities. Another webly supervised learning approach is proposed by Yeung 
\etal \cite{Yeung2017learning}. Here the idea is to learn a data labeling policy 
based on a small set of labeled class instances and to use such a pretrained 
system to gather larger amounts of new training data as well as to label new 
classes based on known concepts.   

\section{Mining actions without annotation}

The here presented dataset is based on the idea that it should be possible to 
learn action classes and concepts from unannotated videos. To do so, we follow 
the idea of \cite{malmaud15what,laptev08learning,bojanowski2014weakly} and make 
use of the spoken language within in the videos to extract possible class 
labels. It follows the idea that, especially in the context of instructional videos, 
people usually explain and comment their actions to the viewers, so the 
performed actions are named during execution, and that those indicators will be 
sufficient to learn action classes from video data without the need for human 
intervention.

\subsection{Test data annotation}
To get the raw material for our system we use freely available video content 
from YouTube. To avoid crawling millions of videos at random we restrict the 
dataset domain to cooking videos and further focus on five simple dishes based 
on egg preparation which are ``eggroll'', ``fried egg'', ``pancake'', 
``omelette'' and ``scrambled egg'', as they all share common tasks and are 
suitable for the mining of related action classes. 
We generate different search queries by combining the five basic recipes with 
standard pre and post fixes to search terms such as ``blueberry pancake'' or 
``easy scrambled egg'' and randomly select 250 videos for our test set, 50 from 
the search query of each dish. 
The test videos are densely annotated, thus each frame is labeled with its 
respective action class.
To support the annotators and to avoid inconsistencies, but still keeping the 
annotation realistic, we prepared a list with ~900 verb object combinations 
which might refer to visible activities. We mined the list automatically from 
textually available online recipes \footnote{recipes gathered from \url{http://www.foodnetwork.com} and \url{http://allrecipes.com}} by applying NLP sentence parsing \cite{Marneffe06generatingtyped} to the related text and keeping all verbs with depended objects. 
The annotators were instructed to label all visible actions in the videos from 
this vocabulary. Additionally, they were also supposed to listen to the 
descriptions from audio comment and give the mentioned verb object combinations 
priority. If an action is visible without audio comment, they were instructed to 
annotate the respective class from the presented list.
Overall we found that the incorporation of the audio comment is crucial because 
of several reasons.  First, if the audio stream is not available, annotators 
class labels do not always match spoken explanation of the respective actions, 
which makes it difficult to evaluate the test set with respect to subtitles. 
Second, giving priority to the audio information helps to avoid an annotator 
bias and to get a representative class distribution within the test data. 
Without audio, annotators usually made use of only a subset of all available 
classes.
All annotated action classes are further checked if they appear at least 10 
times in the training data. Overall, we identify 512 different classes, based on 
94 different verbs and 171 objects. 
\begin{figure}[tb]
    \centering
    \includegraphics[width=0.45\textwidth]{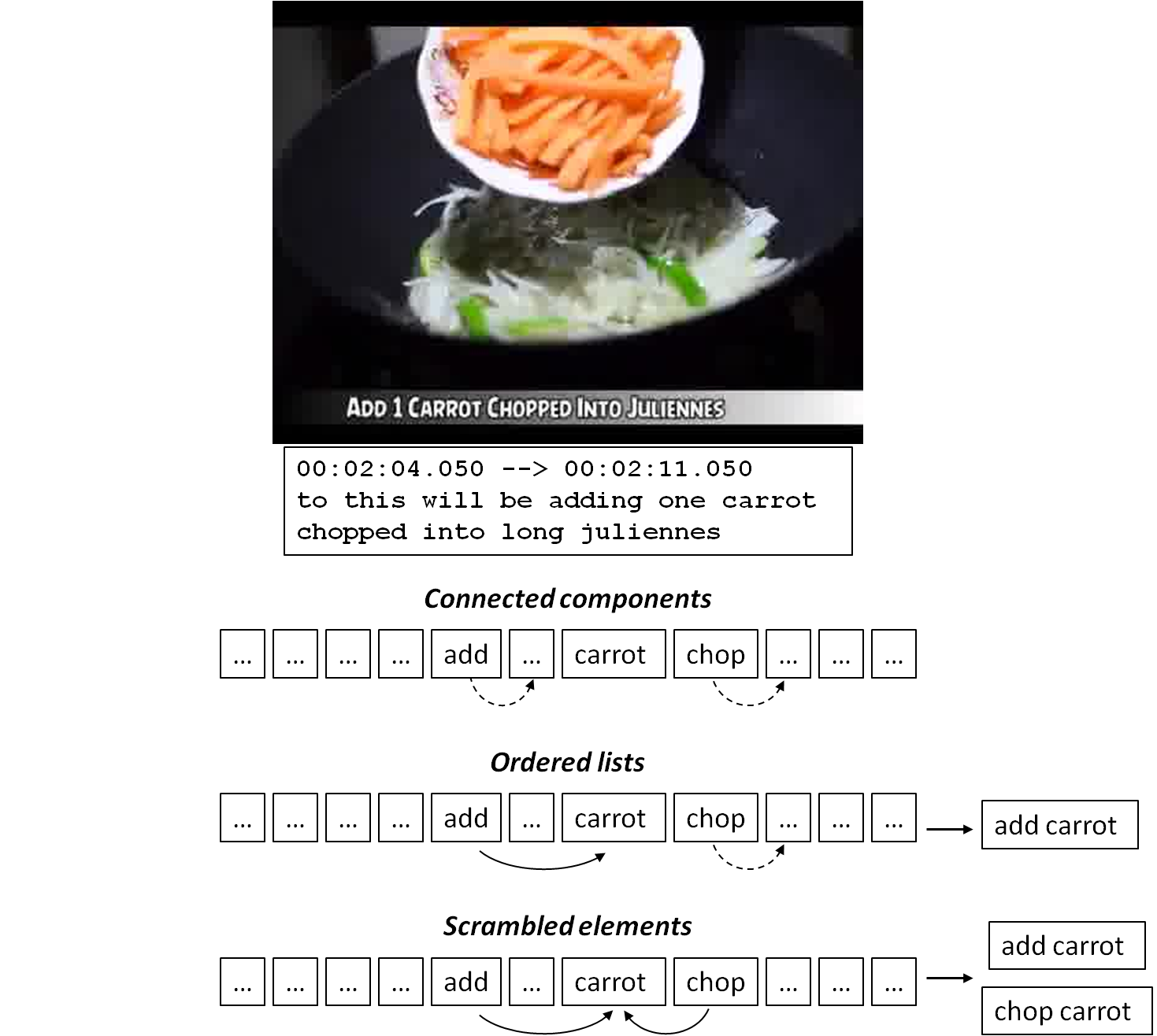}
    \caption{Examples for different mining strategies: (a) In case of mining 
neighboured components, only keywords following directly after another are 
considered as possible class labels. (b) In case of ordered lists, all keyword 
connections from one keyword to the next one are allowed, as long as the order 
is preserved. (c) In case of scrambling, all combinations of all keywords within 
the  subtitle are allowed.  }
    \label{fig:mining_examples}
\end{figure}

\subsection{Mining training data}

To gather the training data we start with the annotated class set and run 
various search modalitites. First, we search for the recipes of respective 
class, e.g. ``recipe crack egg'' as well as for the objects only, e.g. ``recipe 
egg''.  In a second round, we use the titles of the retrieved videos as new 
search queries. We repeat the process with textual recipes as well and use the 
returning titles as another query for our video search. For each search query we 
only use the top 50 results to avoid too much noise in the training data.
This results in 130K video URLs. As many videos are often listed multiple times 
we further pruned the list to only unique entries.

To receive a textual representation of the video content, we lend on the 
close-captioning function of YouTube. The platform provides tools for the owner 
of a video to add closed-captions to the video stream, and, since 2009 also 
provides the option to add automatic generated captions based on Google ASR 
system to the video.

It shows that from the list of possible hits 36,399 unique videos with 
respective closed-captions are available for download, 5842 of them with 
manually added captions and 30557 videos with automatically generated captions. 
One has to note that, although ASR systems show already good performance, there 
is still a noticeable gap between manually edited and automatically generated 
captions. First, there are currently no punctuation marks within the 
automatically generated captions. This is a critical point as they are needed to 
apply NLP parsing to identify subject, verb and dependent object constructs 
within a sentence. Without predefined sentences, there is so far no direct 
off-the-shelf method to acquire this information from running text. Second, 
human annotators usually generate one or two subtitle annotations per sentence 
so that the start and end of subtitles correspond to the beginning and end of 
sentences in spoken language. As we used the start and end time of the 
respective subtitle to generate the class samples, it can in this case be 
assumed that the action is performed within the range of the description. 
Automatic subtitles are based on ongoing speech and are thus always generated for 
fixed temporal segments, usually five to seven seconds. This leads to the fact 
that longer sentences can spread over multiple temporal annotations, which 
increases the chance to miss an action because it is mentioned before the actual 
execution. One further has to remark that in this case it is also not possible 
to infer different action lengths because of fixed temporal segments.
 \begin{table}[t]  \footnotesize
   \centering
   \begin{tabularx}{0.45\textwidth}{lXcXcXc}
        \toprule
        & \multicolumn{6}{c}{Mining criterion}\\
        \cmidrule(lr){1-7}           
        & & Neighbor  & &  Order. & &  Scram. \\
        \cmidrule(lr){1-7}           
        \textit{F. detected} & &   $ 411k  $ & &  $ 1.07M  $ & & $ 1.56M $  
\\
        \textit{F. correct}  & &  $  28k $ & & $  62k $ & & $  89k $  \\
        \textit{Jacc. IoD}  & &  $  3.16 $ & & $  5.44 $& & $  8.35 $  \\
        \textit{Jacc. IoU}  & &  $  1.57 $ & & $  2.58 $ & & $  3.90 $  \\
        \textit{F. hitrate}  & &  $  6.93\% $ & & $  5.85\% $ & & $  
5.68\% $  \\
        \bottomrule
    \end{tabularx}
    \vspace{2mm}
    \caption{Quality of different mining techniques evaluated on the test set. 
Frame-based hitrate describes the amount of videos where at least one frame is 
correctly labeled. } 
    \label{tab:mining_techniques}
\end{table}
\subsection{Subtitle parsing}
Given the list of target classes, we still need to mine respective instances 
from the subtitles and respective video data. To this end, we parse the given 
subtitles for appearances of the target classes, as e.g. the term ``crack egg'' 
in the sentence ``Now we crack three eggs ...''. As in case of automatic 
captions we do not have any punctuation marks and can not rely on sentence 
parsing, we mainly focus on the analysis of word orderings. We evaluate three 
different criteria as shown in Figure \ref{fig:mining_examples}. First, only 
subtitle elements where the verb is directly followed by the object, so only 
direct neighbors, are accepted as class instances. In the second case, we 
consider instances based on ordered lists, thus the verb has to be followed by 
the object, but not directly and third, we mine based on scrambled elements, 
were only the verb and the object have to appear in the respective subtitle 
without any ordering constrains. 
Especially in the last case multiple class labels can be extracted for a single 
clip sample. In this case, we use all detected classes separately. 

To get an idea of the performance of the different mining strategies, we apply 
them to the captions of the test set, which are mainly automatically generated, 
and evaluate the returned samples with respect to the annotated ground truth as 
shown in Table \ref{tab:mining_techniques}. We regard the overall amount of 
detected frames, the amount of correctly detected frames, the Jaccard index of 
the mined samples, computed as intersection over detection and over union, and 
the hitrate which represents the amount of videos in which at least one frame is 
correctly labeled. As can be seen, the scrambling based mining not only produces 
the largest number of training instances, but also the quantitative and relative 
amount of correctly detected frames is larger than for the two other strategies.
 \begin{table}[t] \footnotesize
   \centering
   \begin{tabularx}{0.45\textwidth}{lXcXc}
        \toprule
        & \multicolumn{4}{c}{Comparison of training data}\\
        \cmidrule(lr){1-5}           
        & & Edited & & Automatic \\
        \cmidrule(lr){1-5}           
        \textit{Source Vids}              & &  $ 5842 $ & & $ 30557 $  \\
        \textit{Mined inst.}            & & $  65k  $ & & $ 125k $  \\
        \textit{Mined frames}               & &  $ 10M $ & & $ 23M $  \\
        \textit{Hitrate by anno.}  & & $  46.2\% $ & & $  37.9\% $  
\\
        \bottomrule
    \end{tabularx}
    \vspace{2mm}
    \caption{Quality of different mining techniques evaluated on the test set. 
Hitrate with human judgment describes the amount of videos which were considered 
as correctly labeled by a human annotator i.e. the labeled action was present 
in 
the video. } 
    \label{tab:comp_edited_automatic}
\end{table}
Based on those findings, we apply the scrambling based mining strategy to the 
training set and extract 288k action instances resulting in 52M frames for 
training. 
As this set includes both hand edited and automatically generated subtitles, we 
also wanted to assess how both sets contribute to the overall dataset. 
Therefore, we further analyzed the properties of the  \textit{edited} as well as 
the \textit{automatic} training data separately. An overview of the numbers of 
both data parts is given in Table \ref{tab:comp_edited_automatic}.  One can see 
that the relation of mined instances compared to the number of source videos is 
much higher for the edited subtitles, which mainly goes back to the improved 
quality and temporal alignment of this type of data, but it also shows that 
those samples only make up for one third of the available data. 

To further get an idea about the quality of the training data we used human 
annotators to evaluate the hit rate for both sets. In this case, annotators 
should only decide, if the action label of the respective clip appears somewhere 
in the extracted video clip. We evaluated 2000 videos of each set and refer to 
this as hitrate by human judgment. It shows that the difference between 
the edited and the automatically generated subtitles is only $8.3\%$ with 
$46.2\%$ accuracy in case of edited subtitles and $37.9\%$ for the automatically 
generated data.  
Note that this number only refers to the number of clips in which the labeled 
action was present. But usually not all frames correspond to the labeled 
activity.

To extract meaningful background segments, we further distinguish between 
subtitle annotations, in which objects of verbs from the dictionary appear, but 
without generating an action class, and neutral subtitles without any keywords. 
Only the later ones are considered as background samples. We include $10\%$ of 
such background samples in the training set.

\subsection{Discussion}
We observe that the automatic mining also results in new challenges that do not 
arise in curated datasets with hand selected actions.

First,  as in every real world dataset, we observe a highly imbalanced 
long tail distribution. Overall $102$ classes appear less than $50$ times and 
$187$ classes appear less than 100 times whereas $65$ classes have more than 
$1000$ samples. Additionally, the ten most occurring actions excluding 
background already $25.54\%$ of the training data. One can imaging that 
especially the high ratio of classes with only few samples will make this 
dataset also a valuable source for approaches dealing with the learning from few 
samples, such as one-shot or even zero-shot learning.

Additionally, we found that a lot of automatically mined labels contain semantic inconsistencies. This can be visual
homonyms, thus same labels refer to visually different concepts which can be caused by different object states, \eg egg can refer to a whole egg, thus a white round object, the liquid egg, which can be poured or mixed and the end 
product after preparation \eg in form of scrambled or fried egg. But also regular linguistic polysemies come into play, \eg the word ``pepper'' can refer to spice as well to the vegetable. Complementary, we have linguistic synonyms referring to the same concept by using different words. The most frequent one here is probably the usage of ``add'' and ``pour'', \eg ``add oil'' and ``pour oil'' refer to the same action, but with different linguistic descriptions. This also holds for collective nouns such as ``add spice'' and ``add cumin'' and visually similar objects or actions such as flour and starch.

\section{Hierarchical modeling}
\begin{figure}[t]
    \centering
    \includegraphics[width=0.45\textwidth]{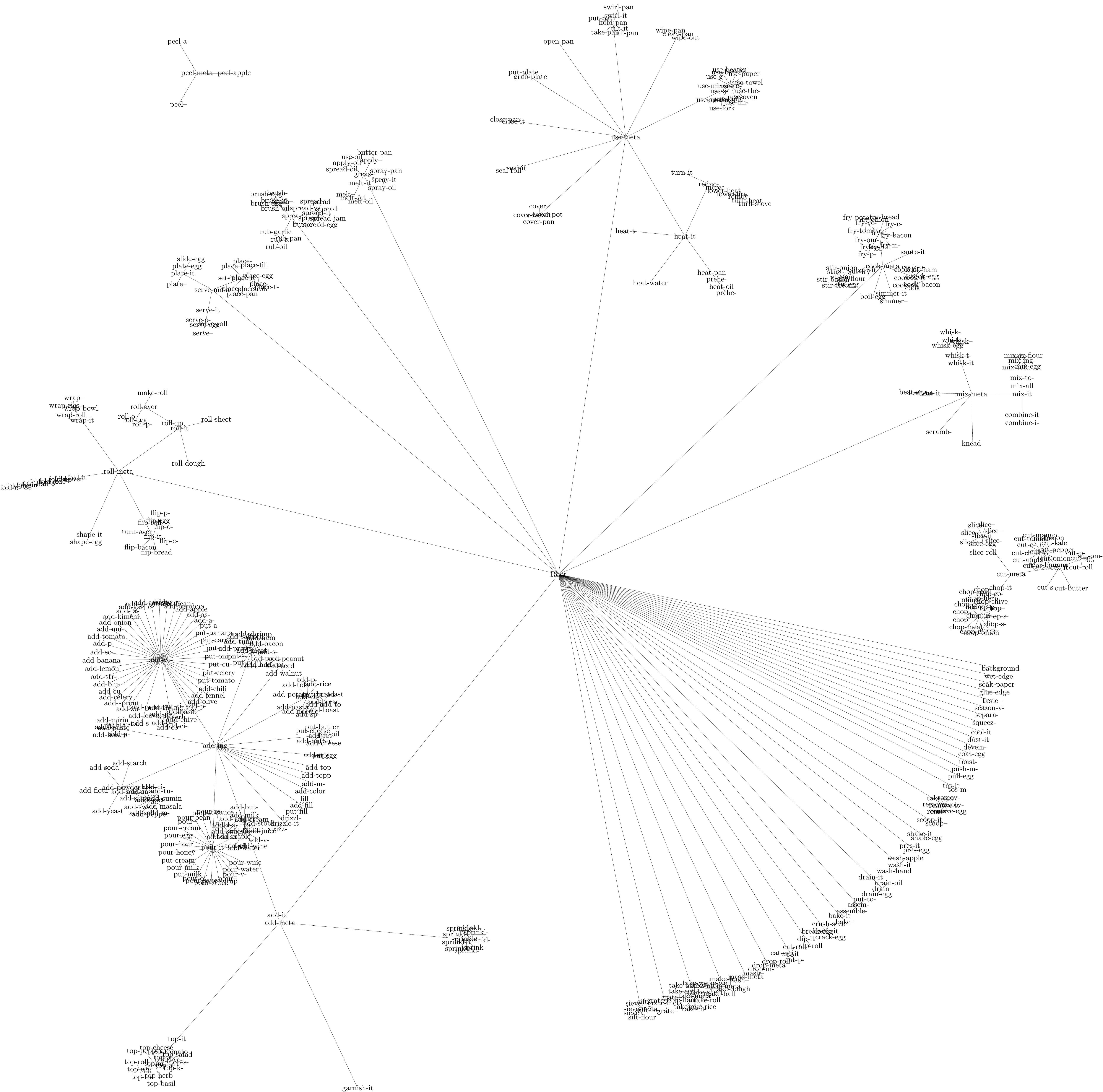}
    \caption{Overview of the hierarchical model for the proposed action classes. }
    \label{fig:hierachy_overview}
\end{figure}
One way to address the problem of semantic inconsistencies, is to model the relatedness of different classes. To give an example for such a relation we can e.g. consider classes such as ``fry it'', ``fry vegetable'', and ``fry tomato''.
They are all referring to a task, but at different levels of abstraction, the first without specifying any object, the second with a general object description, and the last with the most specific one. To capture this inherent structure of the mined actions, we build a hierarchical model of all action classes in the dataset. An overview of the complete structure is given in Figure \ref{fig:hierachy_overview}. 

\begin{figure}[t]
    \includegraphics[width=0.48\textwidth]{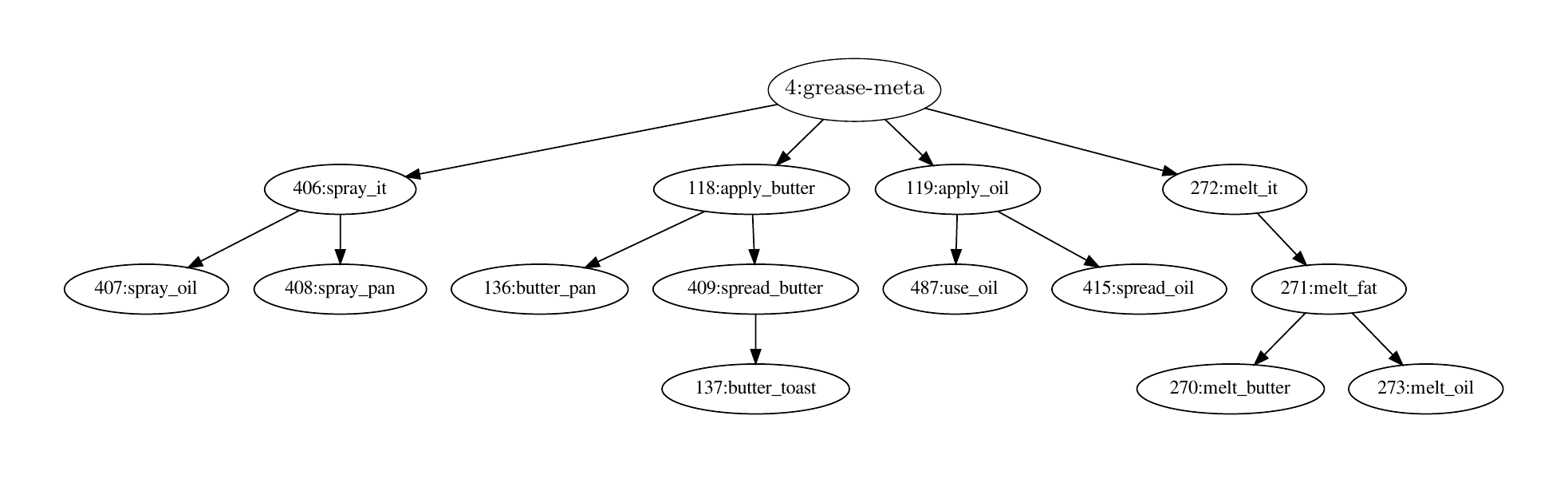}
    \caption{Overview of the hierarchical model for the proposed action classes. }
    \label{fig:graph_example_grease}
\end{figure}

We follow the example of ImageNet \cite{imagenet_cvpr09} that organizes all actions that correspond to more abstract concepts closer to the root while nodes closer to leaves correspond to more detailed labels. Overall, all actions, that are accompanied by a general object such as ``mix it'' are closer to the root, whereas very specific concepts like ``cut basil'' are found at leave nodes. To allow for a consistent hierarchy we further introduce $20$ meta classes ``cut meta'' which serve as a parent class for underlying classes of similar granularity such as ``cut it'', ``chop it'', and ``slice it''. An example of the resulting structure is given in Figure \ref{fig:graph_example_grease} for the branch ``grease meta''. As can be seen from the example, we first consider the verbs to form initial branches in the first layers and then use the related object abstraction granularity to branch out to the leaves. We formulate our model as a hierarchical tree, i.e. each node has only zero or one parent node and can have multiple child nodes with a maximum depth of $18$ levels. 
Note that the hierarchical modeling is also related to a modeling of the overall imbalancing of the dataset, as nodes close to the root usually tend to have more samples because they describe higher-level concepts, whereas classes closer to the leaves usually have less samples, as they describe rather specific tasks such as ``place tortilla'' or ``slice celery'' with only few mined instances.

\section{Evaluation}

\subsection{Feature computation}
To extract features we use the Temporal Segment Network framework as proposed by 
\cite{wang2016temporal}. We used the models pretrained on the Kinetics dataset 
as provided on the website of the authors, as they provide a good trade off 
between accuracy and efficiency. As we do not have any reliable training data, 
we omit fine tuning on a specific set of classes.
We extract features from the output of the last global pooling layer of the 
spatial and temporal stream of the BNInception architecture. We choose the 
architecture as a representative for many similar two stream architectures, as 
it shows that it is able to reach state of the art results on various datasets (see e.g. \cite{carreira2017quo}). To extract frame wise features, we forward only the center crop of each frame through the spatial as well as through the temporal stream. To reduce the overall amount of data, we process every second frame of all clips.

\subsection{Task description}
To evaluate the performance of the different methods and descriptors, we regard the task of temporal alignment of frames to a given ordered set of action classes. This task has been introduced by \cite{bojanowski2014weakly} and has so far been used for most weak learning evaluations \eg \cite{huang2016connectionist,kuehne2016weakly,richard2017weakly, ding2017weakly, richard2018nnviterbi}. 
For this task, the transcript of a video as well as the video itself is available at test time and the goal is to temporally align the video frames to the given list of ordered actions. 
With regard to the proposed dataset, the task of temporal alignment has the advantage that the action classes to detect are already predefined. This relaxes the problem of false synonym labels as in this case \eg ``add oil'' will usually not be mixed up with ``pour oil'' and thus does not result in a false label.
As performance measure, we use the Jaccard index computed as intersection over union (IoU) as well as intersection over detection (IoD) as proposed by \cite{bojanowski2014weakly} as the intersection of ground-truth action $G$ and prediction $D$ as $\frac{G \cap D}{D}$.  We compute both measurements for each video separately and report the mean over all videos.  
Note that IoD in this case only depends of the overall detected segments which leads to better result when one class, e.g. ``background'' becomes larger with respect to all other classes. We therefore mainly consider the standard IoU for the following evaluation and report IoD only as an additional measurement.
 \begin{table}[t] \footnotesize
   \centering
   \begin{tabularx}{0.45\textwidth}{lXcXcXc}
        \toprule
        & \multicolumn{6}{c}{Baseline}\\
        \cmidrule(lr){1-7}           
        & & Linear  & & Random & &Subtitle \\
        \cmidrule(lr){1-7}           
        \textit{Jacc. IoU}  & &  $ 5.45   $ & & $  5.27  $ & & $ 3.90  $  \\
        \textit{Jacc. IoD}  & &  $ 9.12   $ & & $  9.11  $ & & $ 8.35  $  \\
        \bottomrule
    \end{tabularx}
    \vspace{2mm}
    \caption{Results of basic alignment techniques by uniform splitting, random 
splitting as well as for the case of using subtitles only. } 
    \label{tab:eval_random}
\end{table}

\subsection{Baselines}

We first calculate three na\"{i}ve baselines for this task. Here, we consider two setups based on random and uniform alignment, \textit{random guessing} and \textit{uniform distribution} as well as the simple annotation by subtitles only. In case of a temporal alignment task, we know the number of action classes as well as their ordering at test time. For the linear alignment, we simply split the video even according to the number of action segments. In case of random guessing, we assume that frame boundaries are randomly set. We run this setup ten times and receive a mean Jaccard index of $9.11$ for intersection over detection and  $5.27$ for intersection over union. For the uniform distribution we distribute all segments with uniform length over the video. The Jaccard index here is $9.12$ for intersection over detection and  $5.45$ for intersection over union. 

Additionally, we assess the accuracy if we consider labels generated by subtitles only. To this end, we parse the subtitles of the test files for respective appearances of keywords. Note that this evaluation corresponds to an alignment based only on textual input without further training. Here, the Jaccard index is $3.90$ for intersection over union and $8.35$ for intersection over detection.

\subsection{Model evaluation}
\label{sec:model_evaluation}

\begin{table}[t]  \footnotesize
 \centering
  \begin{tabularx}{0.45\textwidth}{lXcXc}
        \toprule
        & \multicolumn{4}{c}{MLP vs GRU}\\
        \cmidrule(lr){1-5}           
          & & MLP  & &  GRU  \\
        \cmidrule(lr){1-5}           
        \textit{Jacc. IoU}  & &  $  9.74  $ & &  $ 7.76 $   \\
        \textit{Jacc. IoD}  & &  $  19.23  $ & &  $ 14.45  $   \\
        \bottomrule
     \end{tabularx}
    \vspace{2mm}
    \caption{Performance of different model architectures measured by the Jaccard index as intersection over detection and over union. } 
    \label{tab:eval_MLP_GRU}
\end{table}

We first regard the performance of the overall setup. To keep the computational load feasible, we run all following experiments with a subset of ~100k samples. To this end, we consider two  network architectures, a multi layer perceptron (MLP) and a network with one layer of gated recurrent units (GRUs) as proposed by \cite{chung2014Empirical}. For the GRUs, we limit the input length to chunks of 10 frames to allow for efficient training. 

Both networks are trained for 15 epochs starting with a learning rate of $ 0.1 $ and reducing it by a factor of $ 0.1 $ every five epochs. At test time, the output probabilities are then transformed into a class-conditional probabilities
$ p(x|c) \propto p(c|x)/p(c) $ using Bayes' rule with $p(c)$ computed from the distribution of classes in the training data. The result is used to align the video frames to the provided transcripts by a Viterbi-decoding as \eg used in~\cite{richard2017weakly}.

For the decoding, we hypothesize segment boundaries over every $ 30 $ frames and restrict the maximum length per action to $1000$ frames. We report results for the best performing system with 2048 units. 

The results for both architectures are shown in Table~\ref{tab:eval_MLP_GRU}. We can see that the GRU is performing lower than the MLP architecture. Looking at the behavior during training, we found that the higher complexity of the GRU starts to overfit the training data at an early stage, which reduces the overall classification accuracy in case of the here used highly-noisy data. 

\begin{table}[t] \footnotesize
    \centering
   \begin{tabularx}{0.45\textwidth}{lXcXcXc}
        \toprule
        & \multicolumn{6}{c}{RGB vs. Optical Flow}\\
        \cmidrule(lr){1-7}           
        & & RGB  & & Flow  && Both\\
        \cmidrule(lr){1-7}           
        \textit{Jacc. IoU}  & & $ 5.87    $ & & $  7.67  $ & & $ 9.74  $ 
\\
        \textit{Jacc. IoD}  & & $ 10.66   $ & & $  14.08  $ & & $ 19.23 $ 
\\
        \bottomrule
    \end{tabularx}
    \vspace{2mm}
    \caption{Results for spatial and temporal flow as well as for the 
combination of both. } 
    \label{tab:eval_rgb_of}
\end{table}

\subsection{Comparison of spatial and temporal stream}

To further analyze the impact of appearance and motion, we evaluate the performance on the spatial and temporal stream separately. As can be seen in Table~\ref{tab:eval_rgb_of}, the single streams alone perform only slightly better than the random baseline. This shows that especially for the case of the here used actions both elements can play a crucial role in the recognition as some classes are rather appearance driven such as ``cut tomato'' but others such as ``whisk egg'' might be more defined by the executed action. 
\subsection{Comparison to webly mined video data}

Another point raised by the proposed approach of mining samples from subtitles 
is the question if the same task could also be achieved by simply using web-crawled videos for the respective classes. To asses the difference of both methods, we used the videos retrieved by the class based search query and trained the model with the respective frame-based features. Here, instead of using mined snippets features from the whole video are considered and labeled by the respective search term. Looking at the results in Table~\ref{tab:eval_webly}, it becomes clear that a vanilla webly supervised training procedure only is performing much worse than a training based on mined instances from subtitles. 

\begin{table}[t] \footnotesize
    \centering
   \begin{tabularx}{0.45\textwidth}{lXcXc}
        \toprule
        & \multicolumn{4}{c}{Webly mined data}\\
        \cmidrule(lr){1-5}           
        & & Web-crawled  & & Subtitle \\
        \cmidrule(lr){1-5}           
        \textit{Jacc. IoU}  & & $ 4.38 $ & & $ 9.74   $  \\
        \textit{Jacc. IoD}  & & $ 9.50 $ & & $ 19.23  $  \\
        \bottomrule
    \end{tabularx}
    \vspace{2mm}
    \caption{Results for the training with web-crawled videos compared to 
samples based on subtitle mining.  } 
    \label{tab:eval_webly}
\end{table}

\subsection{Comparison of subset vs large-scale data }

For efficiency reasons, experiments were so far conducted on a subset of 100k samples. As one of the main arguments for an automatic mining of training samples, we now assess how far an increase of training samples influences the overall performance of the system. To this end, we train the same configuration with the full amount of 200k samples. As can be seen in Table \ref{tab:eval_large_scale}, doubling the amount of training samples leads to an increase more than $2\%$ with respect to IoU.
Thus, adding more training samples helps, even if a large amount of noise is included, but it also shows that the number of training samples especially for rare classes is still not enough to significantly increase the overall performance of the system. 

\begin{table}[t] \footnotesize
 \centering
  \begin{tabularx}{0.45\textwidth}{lXcXc}
        \toprule
        & \multicolumn{4}{c}{Subset vs large-scale data}\\
        \cmidrule(lr){1-5}           
         samples & &  $ 100k $ & &  $ 200k $  \\
        \cmidrule(lr){1-5}           
        \textit{MLP (IoU)}  & &  $  9.14  $ & &  $ 11.42  $   \\
        \textit{MLP (IoD)}  & &  $  19.23  $ & &  $ 18.79  $   \\
        \bottomrule
     \end{tabularx}
    \vspace{2mm}
    \caption{Performance of training with 100k compared to 200k samples. } 
    \label{tab:eval_large_scale}
\end{table}

\subsection{Hierarchical inference}

 \begin{table}[t]  \footnotesize
   \centering
   \begin{tabularx}{0.45\textwidth}{lXcXcXc}
        \toprule
        & \multicolumn{6}{c}{Hierachical inference - fully}\\
        \cmidrule(lr){1-7}           
         & & bottom-up & & top-down  & & t+b \\
        \cmidrule(lr){1-7}           
        \textit{Jacc. IoU}  & &  $ 14.02  $ & & $ 12.62  $ & & $ 14.81 $  \\
        \textit{Jacc. IoD}  & &  $ 23.57  $ & & $ 26.03  $ & & $ 30.83 $  \\
        \bottomrule
    \end{tabularx}
   \vspace{2mm}
   \begin{tabularx}{0.45\textwidth}{lXcXcXc}
        \toprule
        & \multicolumn{6}{c}{Hierachical inference - pairwise}\\
        \cmidrule(lr){1-7}           
         & & bottom-up & & top-down  & & t+b \\
        \cmidrule(lr){1-7}           
        \textit{Jacc. IoU}   & &  $ 14.11 $ & & $  11.62 $ & & $ 15.00 $  \\
        \textit{Jacc. IoD}   & &  $ 22.99 $ & & $  19.86 $ & & $ 26.83  $  \\
        \bottomrule
    \end{tabularx}
    \vspace{2mm}
    \caption{Comparison of hierarchical inference strategies of the full dataset with 200k samples. } 
    \label{tab:eval_hier_inf_strat}
\end{table}
Finally, we evaluate the impact of the hierarchical model on the proposed architecture. To this end we compute the consensus probability in a hierarchical top-down and bottom-up inference as proposed by \cite{Valentini2011True}. The idea of the consensus probability is that, is a node at a certain level of granularity is true, this also has to hold for its parent nodes. If e.g. in cased of the succeeding nodes, \eg ``flip-it'' $\rightarrow$ ``flip-egg'' $\rightarrow$ ``flip-omlet'', ``flip-egg'' is recognized, than the related parent class ``flip it'' should also show a higher response. The same holds for the bottom up inference, if a high level class is recognized, then there should also be a higher response from its child classes than from classes in other branches. Note that we also tried several approaches to include the hierarchical information during training, but so far none of them was able to outperform the here used consensus voting.  

For top-down inference, we multiply the probabilities of all parents with one of the current node, whereas for the bottom-up approach we consider at each level the mean probability of all child nodes and multiply those with the one of the parent until the current node is reached.  We combine the result of top-down and bottom-up inference by considering the mean of top-down and bottom up inference. We further evaluate the performance of the different consensus for computing top-down and bottom-up probabilities for the full graph as well as for pairwise combination of neighboring nodes only. As Table \ref{tab:eval_hier_inf_strat} shows, both methods of consensus voting are able to improve the overall accuracy with best results obtained by combining both inference strategies.
Using the hierachical inference increases the IoU by over $3.5\%$ and the IoD by over $12\%$.
It can thus be assumed that the consensus probability helps in this setting as it provides a form of regularization for the noisy training data, as in this case it is not enough for a single class to have a high response with respect to some frames, which can also be influenced by of noise in the training data. Instead multiple semantically related classes need to have a similar high response for those frames, which is less likely for random noisy clips than  for the real training class in this case.
\begin{figure}[t]
    \centering
    \includegraphics[width=0.40\textwidth]{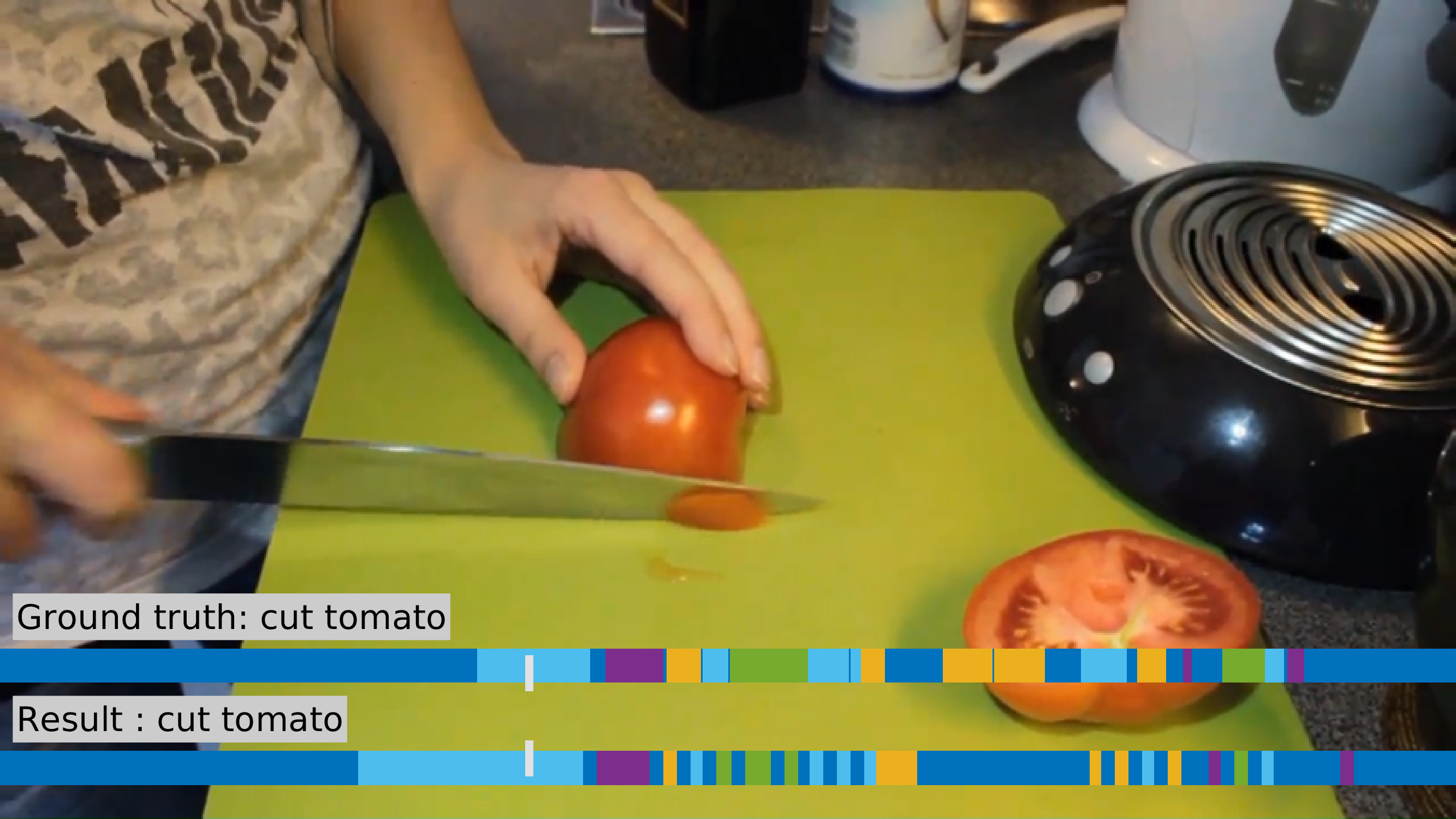}
    \\
    \includegraphics[width=0.40\textwidth]{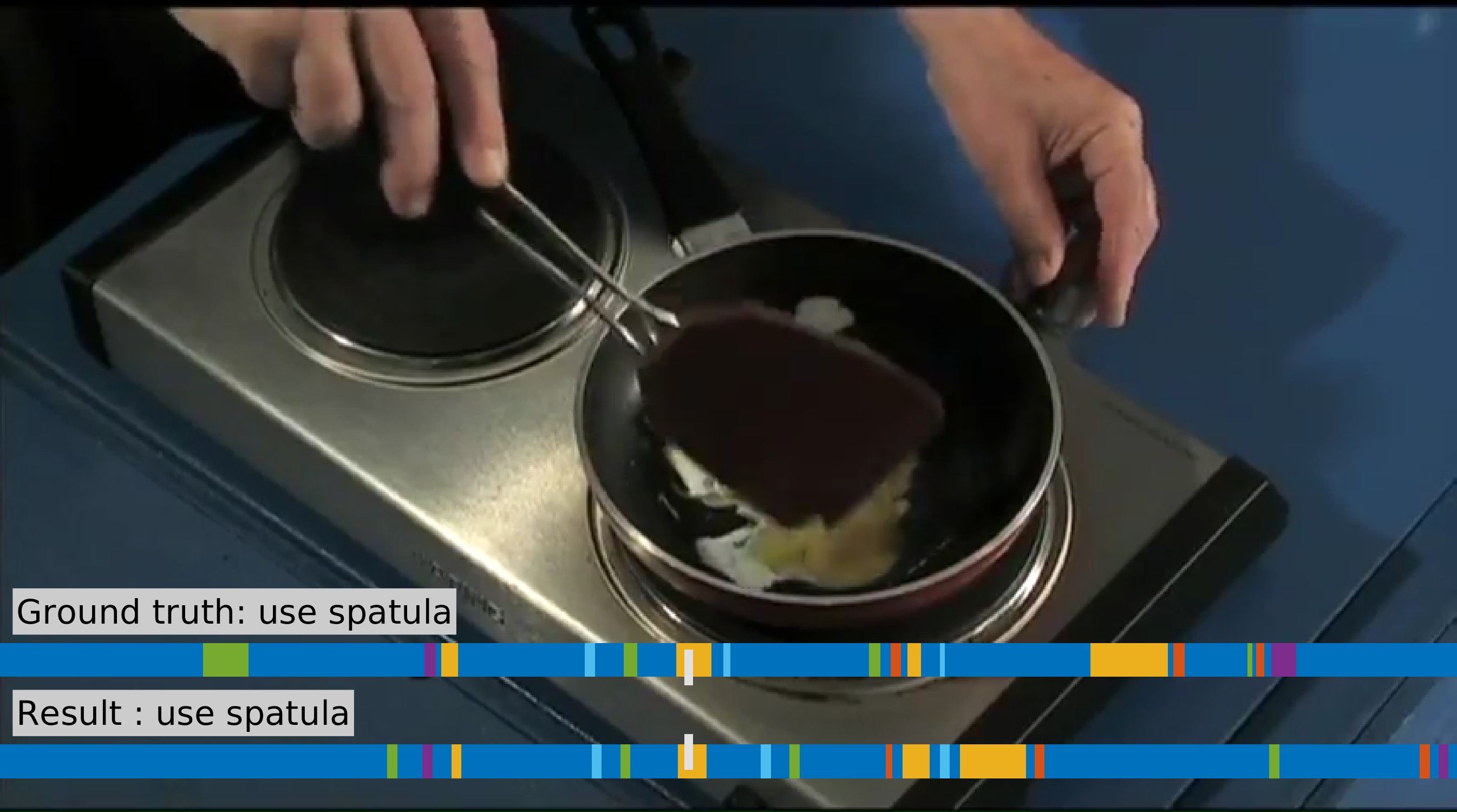}
    \caption{Qualitaive examples of the alignment of two sample videos from the test set. }
    \label{fig:examples_classes}
\end{figure}
%


%
%
\section{Conclusion}

We proposed a new dataset for the challenging task of learning action classes from unannotated videos. Using subtitles, human edited as well as automatically generated ones, we were able to automatically extract action classes from the video stream and to train them without any need for human intervention. We also proposed a hierarchical tree structure showing that the modeling of semantically similar actions at different levels of granularity can help to improve the overall performance significantly in such noisy scenarios.

We think that this setup holds a huge potential for the field of weak action learning as well as for action recognition in general, as it allows the mining of training data at a new scale. Further, the idea of mining training data automatically from subtitles might help to close the gap between nowadays popular hand crafted datasets and real-world applications and training scenarios.

{\small
\bibliographystyle{ieee}
\bibliography{references}

\begin{thebibliography}{10}\itemsep=-1pt

\bibitem{alayrac2016unsupervised}
J.-B. Alayrac, P.~Bojanowski, N.~Agrawal, I.~Laptev, J.~Sivic, and
  S.~Lacoste-Julien.
\newblock Unsupervised learning from narrated instruction videos.
\newblock In {\em IEEE Conf. on Computer Vision and Pattern Recognition}, 2016.

\bibitem{arandjelovic2017looklistenlearn}
R.~Arandjelovic and A.~Zisserman.
\newblock Look, listen and learn.
\newblock {\em arXiv preprint arXiv:1705.08168}, 2017.

\bibitem{blank2005database}
M.~Blank, L.~Gorelick, E.~Shechtman, M.~Irani, and R.~Basri.
\newblock Actions as space-time shapes.
\newblock In {\em Int. Conf. on Computer Vision}, pages 1395--1402, 2005.

\bibitem{bojanowski2014weakly}
P.~Bojanowski, R.~Lajugie, F.~Bach, I.~Laptev, J.~Ponce, C.~Schmid, and
  J.~Sivic.
\newblock Weakly supervised action labeling in videos under ordering
  constraints.
\newblock In {\em European Conf. on Computer Vision}, pages 628--643, 2014.

\bibitem{carreira2017quo}
J.~Carreira and A.~Zisserman.
\newblock Quo vadis, action recognition? {A} new model and the kinetics
  dataset.
\newblock In {\em CVPR}, pages 4724--4733, 2017.

\bibitem{chung2014Empirical}
J.~Chung, C.~Gulcehre, K.~Cho, and Y.~Bengio.
\newblock Empirical evaluation of gated recurrent neural networks on sequence
  modeling.
\newblock 2014.

\bibitem{Marneffe06generatingtyped}
M.-C. de~Marneffe, B.~MacCartney, and C.~D. Manning.
\newblock Generating typed dependency parses from phrase structure parses.
\newblock In {\em Int. Conf. on Language Resources and Evaluation (LREC)},
  pages 449--454, 2006.

\bibitem{imagenet_cvpr09}
J.~Deng, W.~Dong, R.~Socher, L.-J. Li, K.~Li, and L.~Fei-Fei.
\newblock {ImageNet: A Large-Scale Hierarchical Image Database}.
\newblock In {\em CVPR}, 2009.

\bibitem{ding2017weakly}
L.~Ding and C.~Xu.
\newblock Weakly-supervised action segmentation with iterative soft boundary
  assignment.
\newblock In {\em CVPR}, 2017.

\bibitem{caba2015activitynet}
B.~G. Fabian Caba~Heilbron, Victor~Escorcia and J.~C. Niebles.
\newblock Activitynet: A large-scale video benchmark for human activity
  understanding.
\newblock In {\em IEEE Conf. on Computer Vision and Pattern Recognition}, pages
  961--970, 2015.

\bibitem{gu2017ava}
C.~Gu, C.~Sun, S.~Vijayanarasimhan, C.~Pantofaru, D.~A. Ross, G.~Toderici,
  Y.~Li, S.~Ricco, R.~Sukthankar, C.~Schmid, and J.~Malik.
\newblock {AVA:} {A} video dataset of spatio-temporally localized atomic visual
  actions.
\newblock {\em arXiv preprint arXiv:1705.08421}, 2017.

\bibitem{hong2017weakly}
S.~Hong, D.~Yeo, S.~Kwak, H.~Lee, and B.~Han.
\newblock Weakly supervised semantic segmentation using web-crawled videos.
\newblock 2017.

\bibitem{huang2016connectionist}
D.-A. Huang, L.~Fei-Fei, and J.~C. Niebles.
\newblock Connectionist temporal modeling for weakly supervised action
  labeling.
\newblock In {\em European Conf. on Computer Vision}, pages 137--153, 2016.

\bibitem{thumos14}
Y.-G. Jiang, J.~Liu, A.~Roshan~Zamir, G.~Toderici, I.~Laptev, M.~Shah, and
  R.~Sukthankar.
\newblock {THUMOS} challenge: Action recognition with a large number of
  classes, 2014.

\bibitem{kaufman2016tessellation}
D.~Kaufman, G.~Levi, T.~Hassner, and L.~Wolf.
\newblock Temporal tessellation for video annotation and summarization.
\newblock {\em arXiv preprint arXiv:1612.06950}, 2016.

\bibitem{kuehne2014breakfast}
H.~Kuehne, A.~Arslan, and T.~Serre.
\newblock The language of actions: Recovering the syntax and semantics of
  goal-directed human activities.
\newblock In {\em IEEE Conf. on Computer Vision and Pattern Recognition}, pages
  780--787, 2014.

\bibitem{kuehne2011hmdb}
H.~Kuehne, H.~Jhuang, E.~Garrote, T.~Poggio, and T.~Serre.
\newblock {HMDB}: a large video database for human motion recognition.
\newblock In {\em Int. Conf. on Computer Vision}, pages 2556--2563, 2011.

\bibitem{kuehne2016weakly}
H.~Kuehne, A.~Richard, and J.~Gall.
\newblock Weakly supervised learning of actions from transcripts.
\newblock {\em arXiv preprint arXiv:1610.02237}, 2016.

\bibitem{laptev08learning}
I.~Laptev, M.~Marszalek, C.~Schmid, and B.~Rozenfeld.
\newblock Learning realistic human actions from movies.
\newblock In {\em IEEE Conf. on Computer Vision and Pattern Recognition}, 2008.

\bibitem{malmaud15what}
J.~Malmaud, J.~Huang, V.~Rathod, N.~Johnston, A.~Rabinovich, and K.~Murphy.
\newblock What{'}s cookin{'}? {I}nterpreting cooking videos using text, speech
  and vision.
\newblock In {\em Conf. of the North American Chapter of the Association for
  Computational Linguistics: Human Language Technologies}, 2015.

\bibitem{richard2017weakly}
A.~Richard, H.~Kuehne, and J.~Gall.
\newblock Weakly supervised action learning with rnn based fine-to-coarse
  modeling.
\newblock In {\em IEEE Conf. on Computer Vision and Pattern Recognition}, 2017.

\bibitem{richard2018nnviterbi}
A.~Richard, H.~Kuehne, A.~Iqbal, and J.~Gall.
\newblock Neuralnetwork-viterbi: A framework for weakly supervised video
  learning.
\newblock In {\em CVPR}, 2018.

\bibitem{rohrbach2012database}
M.~Rohrbach, S.~Amin, M.~Andriluka, and B.~Schiele.
\newblock A database for fine grained activity detection of cooking activities.
\newblock In {\em IEEE Conf. on Computer Vision and Pattern Recognition}, pages
  1194--1201, 2012.

\bibitem{kth2004database}
C.~Schuldt, I.~Laptev, and B.~Caputo.
\newblock Recognizing human actions: a local svm approach.
\newblock In {\em CVPR}, volume~3, pages 32--36, 2004.

\bibitem{Sener2017unsupervised}
O.~Sener, A.~Zamir, S.~Savarese, and A.~Saxena.
\newblock Unsupervised semantic parsing of video collections.
\newblock In {\em ICCV}, 2015.

\bibitem{sigurdsson2016hollywood}
G.~A. Sigurdsson, G.~Varol, X.~Wang, A.~Farhadi, I.~Laptev, and A.~Gupta.
\newblock Hollywood in homes: Crowdsourcing data collection for activity
  understanding.
\newblock In {\em European Conference on Computer Vision}, 2016.

\bibitem{simonyan2014two}
K.~Simonyan and A.~Zisserman.
\newblock Two-stream convolutional networks for action recognition in videos.
\newblock In {\em Advances in Neural Information Processing Systems}, pages
  568--576, 2014.

\bibitem{soomro2012ucf101}
K.~Soomro, A.~R. Zamir, and M.~Shah.
\newblock Ucf101: A dataset of 101 human actions classes from videos in the
  wild.
\newblock {\em arXiv preprint arXiv:1212.0402}, 2012.

\bibitem{Valentini2011True}
G.~Valentini.
\newblock True path rule hierarchical ensembles for genome-wide gene function
  prediction.
\newblock {\em IEEE/ACM Trans. Comput. Biol. Bioinformatics}, 8(3), May 2011.

\bibitem{wang2013action}
H.~Wang and C.~Schmid.
\newblock Action recognition with improved trajectories.
\newblock In {\em Int. Conf. on Computer Vision}, pages 3551--3558, 2013.

\bibitem{wang2016temporal}
L.~Wang, Y.~Xiong, Z.~Wang, Y.~Qiao, D.~Lin, X.~Tang, and L.~{Val Gool}.
\newblock Temporal segment networks: Towards good practices for deep action
  recognition.
\newblock In {\em ECCV}, 2016.

\bibitem{Yeung2017learning}
S.~Yeung, V.~Ramanathan, O.~Russakovsky, L.~Shen, G.~Mori, and L.~Fei-Fei.
\newblock Learning to learn from noisy web videos.
\newblock In {\em CVPR}, 2017.

\bibitem{zhou2017procnets}
L.~Zhou, C.~Xu, and J.~J. Corso.
\newblock Towards automatic learning of procedures from web instructional
  videos.
\newblock {\em arXiv preprint arXiv:1703.09788}, 2017.

\end{thebibliography}
}

\end{document}